\documentclass[runningheads]{llncs}
\usepackage[final]{graphicx}

\usepackage{algorithm}
\usepackage{algorithmic}
\usepackage{url}
\usepackage[table,xcdraw]{xcolor}
\usepackage{subfig}

\begin{document}
\title{APS: A Large-Scale Multi-Modal Indoor Camera Positioning System}
%
%\titlerunning{Abbreviated paper title}
% If the paper title is too long for the running head, you can set
% an abbreviated paper title here
%
\author{Ali Ghofrani\inst{1,2} \and
	Rahil Mahdian Toroghi\inst{1} \and
	Seyed Mojtaba Tabatabaie\inst{2}}
%\orcidID{0000-0002-7668-7369}
%\author{First Author\inst{1}\orcidID{0000-1111-2222-3333} \and
%Second Author\inst{2,3}\orcidID{1111-2222-3333-4444} \and
%Third Author\inst{3}\orcidID{2222--3333-4444-5555}}
%%
%\authorrunning{F. Author et al.}
% First names are abbreviated in the running head.
% If there are more than two authors, 'et al.' is used.
%
\institute{Iran Broadcasting University (IRIBU), Tehran, Iran \and
    CEO/CTO at Alpha Reality, AR/VR Solution Company	
	 \\
	\email{alighofrani@iribu.ac.ir}\\
	\email{mahdian.t.r@gmail.com}\\
%	\email{\{engalighofrani,mahdian.t.r\}@gmail.com}\\
	\email{smtabatabaie@alphareality.io}
}

%\institute{Princeton University, Princeton NJ 08544, USA \and
%Springer Heidelberg, Tiergartenstr. 17, 69121 Heidelberg, Germany
%\email{lncs@springer.com}\\
%\url{http://www.springer.com/gp/computer-science/lncs} \and
%ABC Institute, Rupert-Karls-University Heidelberg, Heidelberg, Germany\\
%\email{\{abc,lncs\}@uni-heidelberg.de}}
%
\maketitle              % typeset the header of the contribution
\vspace{-3mm}
\begin{abstract}
Navigation inside a closed area with no GPS-signal accessibility is a highly challenging task. In order to tackle this problem, recently the imaging-based methods have grabbed the attention of many researchers. These methods either extract the features (e.g. using SIFT, or SOSNet) and map the descriptive ones to the camera position and rotation information, or deploy an end-to-end system that directly estimates this information out of RGB images, similar to PoseNet. While the former methods suffer from heavy computational burden during the test process, the latter suffers from lack of accuracy and robustness against environmental changes and object movements. However, end-to-end systems are quite fast during the test and inference and are pretty qualified for real-world applications, even though their training phase could be longer than the former ones. In this paper, a novel multi-modal end-to-end system for large-scale indoor positioning has been proposed, namely APS (Alpha Positioning System), which integrates a Pix2Pix GAN network to reconstruct the point cloud pair of the input query image, with a deep CNN network in order to robustly estimate the position and rotation information of the camera. For this integration, the existing datasets have the shortcoming of paired RGB/point cloud images for indoor environments. Therefore, we created a new dataset to handle this situation. By implementing the proposed APS system, we could achieve a highly accurate camera positioning with a precision level of less than a centimeter.

\keywords{Indoor camera positioning  \and Point-cloud data \and Convlutional Neural Network \and Pix2Pix GAN.}
\end{abstract}
%
%
%

%\vspace{-3mm}
\section{Introduction}
\vspace{-2mm}
Throughout the development of navigation systems, there are lots of challenges along the way, which have been mainly solved by development of GPS systems. However, the indoor navigation, due to the lack of GPS data, has remained unsolved, yet. Inside malls, airports, large towers, warehouses, and many other places the position can easily get lost, when the GPS data is hardly available. 

One solution has been to place very small bluetooth broadcasting devices all over the area, so that they can detect the target and based on their distribution topology predict the position. However, that could have several disadvantages~\cite{faragher2014analysis,mendoza2019meta,huang2019hybrid,jianyong2014rssi}. For example, extra devices become necessary which entails periodic monitoring and maintenance, as well as calibrations. Moreover, the battery lifetime of the bluetooth devices would be a critical issue and the target could be easily lost when the battery goes off. In addition, only a rough estimation of the position could be achieved and this may hardly work in a crowded and winding corridor. With a same logic, using WiFi signals have also been leveraged ~\cite{yang2015wifi,caso2019vifi,sharp2019indoor}, yet the position precision has not been highly improved and the installation of extra devices would still be a need ~\cite{mekki2019indoor,morgado2019beacons}. The latest Apple localization system using WiFi signals, have achieved the accuracy of 3 to 5 meters~\cite{zuo2018indoor}. Although, this precision might be acceptable for outdoor navigations, it may not work for indoor scenarios, such as autonomous driving, or augmented reality applications in which the digital overlays require a centimeter precision scale~\cite{ghofrani2019icps,shen2019low,sieberth2019applying,russ2019augmented}.

A different solution has been through geometry-aware systems, which incorporate perceptual and temporal features of camera imaging in order to extract the position and quaternion information. This is pretty similar to the human way of memorizing a location and navigating by using visual and temporal features. Some conventional methods in this regard, either use RGB images along with a depth-assisted camera~\cite{zhang2018real,wang2019high,henry2014rgb,yuan2016rgb,duque2017improved,lai2018development}, or they employ SIFT-based algorithms~\cite{sattler2016efficient,valgren2010sift,liang2013image}. In many realistic scenarios however, the depth-based camera or bluetooth or WiFi signals are not available~\cite{duque2016indoor,shao2018ibeacon}, at all.

In this paper, following the works of ~\cite{ghofrani2019icps,ghofrani2019lidar} we present an end-to-end deep neural network system that involves the RGB data of a particular scene in one hand and the point-cloud data corresponding to it on the other hand; then integrates them and provides the camera position and quaternion estimates with a high accuracy. Furthermore, the proposed system is proved to be robust against environmental variations including partially masking of the images, light changes, etc. in contrast to ICPS-net work~\cite{ghofrani2019icps}.
\begin{figure*}[!t]
	\centering
	\includegraphics[width=1\linewidth, height=.21\textheight]{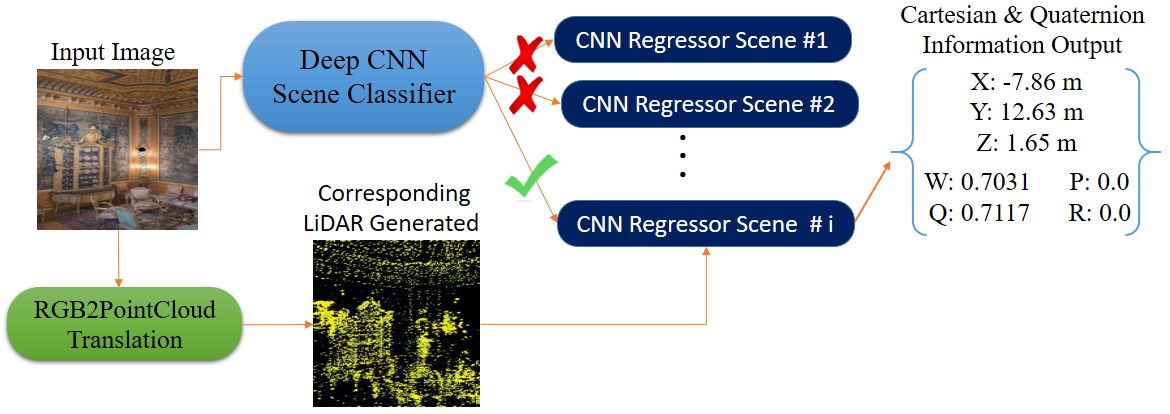}
	\caption{The entire proposed APS process flowgraph, pictorially demonstrated.}
	\label{fig:big_Pic}
	\vspace{-2mm}
\end{figure*}

The outline goes, as follows. First, the related works are explained. Then, a big-picture of the proposed end-to-end system along with its building blocks is introduced. Next, the training and evaluation processes of the architecture are explained. In addition, a dataset has been created for this purpose, and the way to employ it for the training processes is mentioned. Alongside, the experiments and the evaluation measures are introduced. The paper is finally terminated by the results, conclusion, and the references being cited within the contained sections.

%\vspace{-1mm}
\section{Related Works}
\vspace{-2mm}
A pioneering work for outdoor positioning using deep neural networks proposed in~\cite{kendall2015posenet}, as PoseNet. This system employs a convolutional neural network for a real-time 6 degree-of-freedom camera relocalization.
MapNet is another deep network based system, which uses geometrical information of the images for the camera localization~\cite{henriques2018mapnet}. Both of these methods, were presented to address the outdoor positioning problem.

For outdoor places, the positioning is possible due to using image processing methods and the existence of indicative objects in the environment as the markers. However, for indoor places, due to limited scales and conditions, there would be many identical patterns which hinder the SIFT or SURF-based methods to work well in practice~\cite{guan2016vision,liang2015image,valgren2010sift,sattler2012improving,brahmbhatt2018geometry}.
Moreover, inside buildings the configurations are usually subject to changes such as object replacements or movements, which in turn modify the status of the patterns over the course of the imaging period.

In order to pave the way for indoor positioning, a deep conventional neural network system, namely ICPS-net~\cite{ghofrani2019icps} was proposed, primarily. In this work, a photogrammetry was performed on selected scenes in the area of interest. An EfficientNet CNN network~\cite{tan2019efficientnet}, is then employed which has been already trained on the sequence of images associated to each specific scene inside the interested area. This CNN network, then takes in the user image and classifies it as one of the previously trained scenes. When the scene associated to the user input image is identified, another MobileNet-V2 CNN is employed~\cite{sandler2018mobilenetv2}, which has been already trained to perform the regression over the camera Cartesian positions, as well as the quaternion information. While these tandem connected networks as a system is able to achieve a high position accuracy, it suffers from the lack of robustness due to environmental changes that could easily confuse the system. Another 
problem with this system is that creating data for the model requires excruciating efforts. Later, a LiDAR ICPS-net~\cite{ghofrani2019lidar} was proposed which involves the point cloud data being extracted through a LiDAR system to make the model sufficiently robust against the environmental changes. The disadvantage of this model was that it could  get stuck in the model convergence problem, and the training phase could be extremely prolonged. In addition, the accuracy was deteriorated significantly with respect to the ICPS-net system with RGB data. 

In the current work, we are integrating the prominent properties of the ancestral ICPS-net based systems and we integrate them so that the final architecture performs a higher precision estimates while the entire data preparation and model convergence, as well as the estimation precision are all significantly improved. Briefly speaking, a place is initially introduced to our system  for indoor navigation. We divide it into different scenes, and we show all the possible views of the scenes through camera moving strategies. Then, we train our model based on these obtained images for the purpose of scene classification and final navigation step.

%\vspace{-1mm}
\section{The Proposed Architecture}
\begin{figure}[!ht]
	\centering
	\includegraphics[width=.52\linewidth, height=.47\textheight]{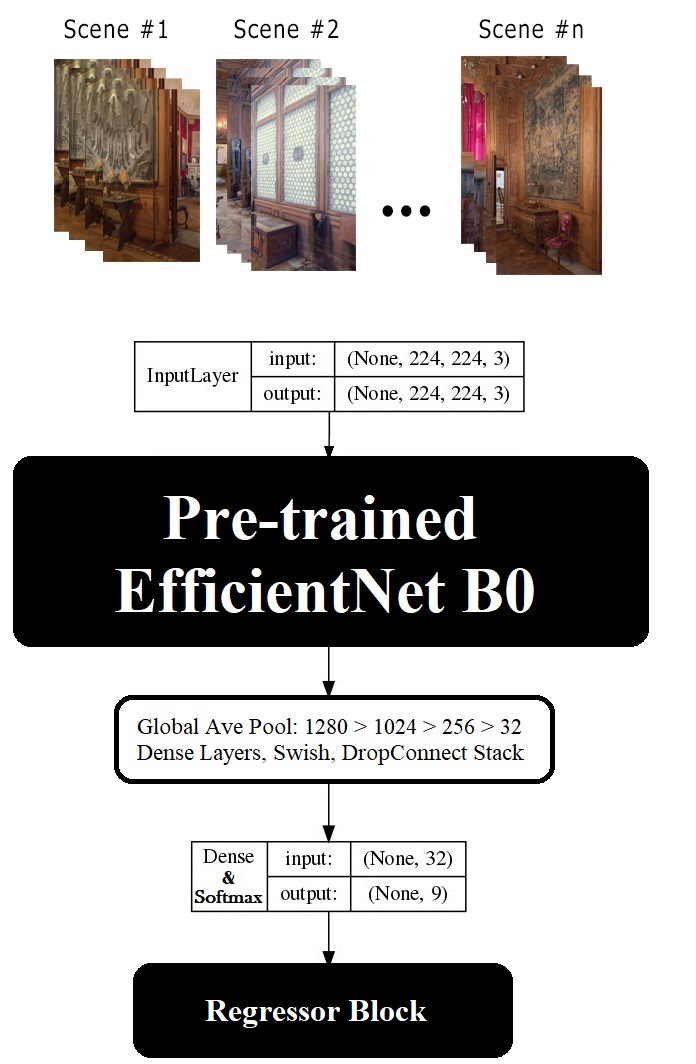}
	\caption{Scene classifier based on EfficientNet-B0 CNN. \textbf{Input}:RGB data; \textbf{Output}:Associated scene number.}
	\label{fig:classifier}
	\vspace{-4mm}
\end{figure}

\vspace{-2mm}
As depicted in the system architecture in figure~\ref{fig:big_Pic}, there is a convolutional neural network (CNN), which is trained in order to identify the specified scene out of divided segments form the interested area based on the user input RGB image taken by the mobile phone,  for instance. The input image size is changed to become compatible with the employed CNN structure (here, an EfficientNet B0 network~\cite{sattler2016efficient}, as in figure~\ref{fig:classifier}). The point-cloud data associated to the RGB images of the interested area have to be already extracted using a LiDAR system. The entire process has been explained in~\cite{ghofrani2019lidar}.  
\vspace{-3mm}

During training of the system, the imagery RGB data is given to a UNET-based network (as in figure~\ref{fig:unet}) to reconstruct the point-cloud image associated to the RGB image~\cite{ronneberger2015u}. This network is a Pix2Pix GAN (\textit{Generative Adversarial Network}), which produces these corresponding pairs of (RGB, Point-cloud) images~\cite{isola2017image}. The input RGB images are augmented for the Pix2Pix GAN network, using different contrast and brightness levels as well as masking objects as depicted in figure~\ref{fig:masked},  in order to make the reconstruction of pointcloud images of the network sufficiently robust. Then, two CNN models are trained, simultaneously on the RGB images, as well as the LiDAR reconstructed point-cloud images, respectively. These CNNs are trained in a parallel regime, to perform as the regression models for the position estimates. 
The complete data (including RGB, Point-cloud and augmented ones) with more than one million samples will be released for further research works.
The process is explained in the sequel.
\subsection{Created Dataset}
\vspace{-10mm}

\begin{figure}[!ht]
	\subfloat[\label{fig:3Dmodel1}]{%
		\includegraphics[width=.49\linewidth, height=.23\textheight]{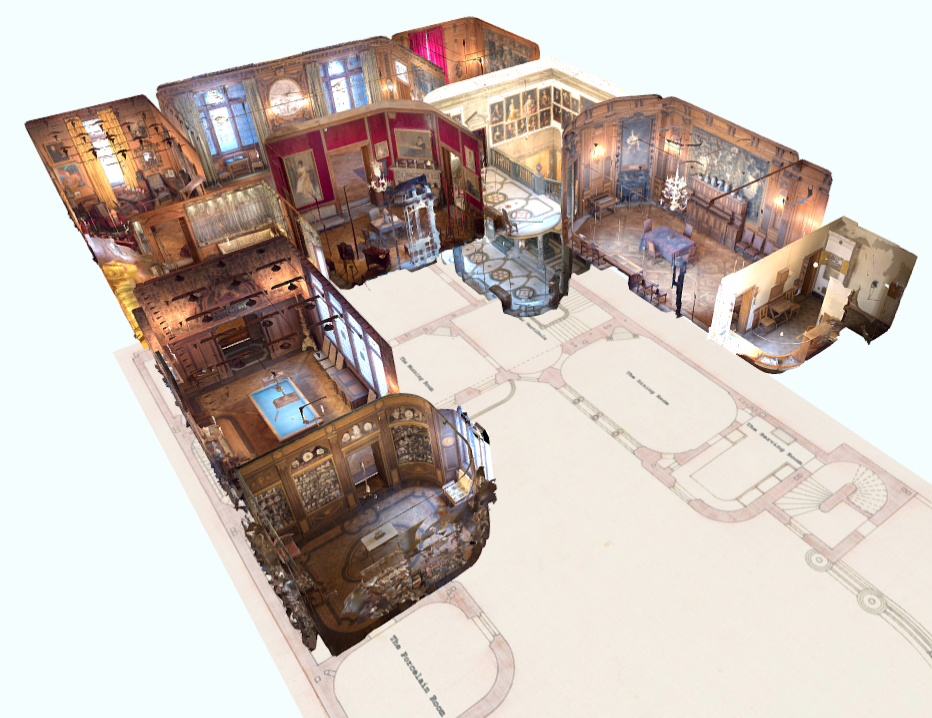}
	}
	\hfill
	\subfloat[\label{fig:3Dmodel_PC}]{%
		\includegraphics[width=.49\linewidth, height=.23\textheight]{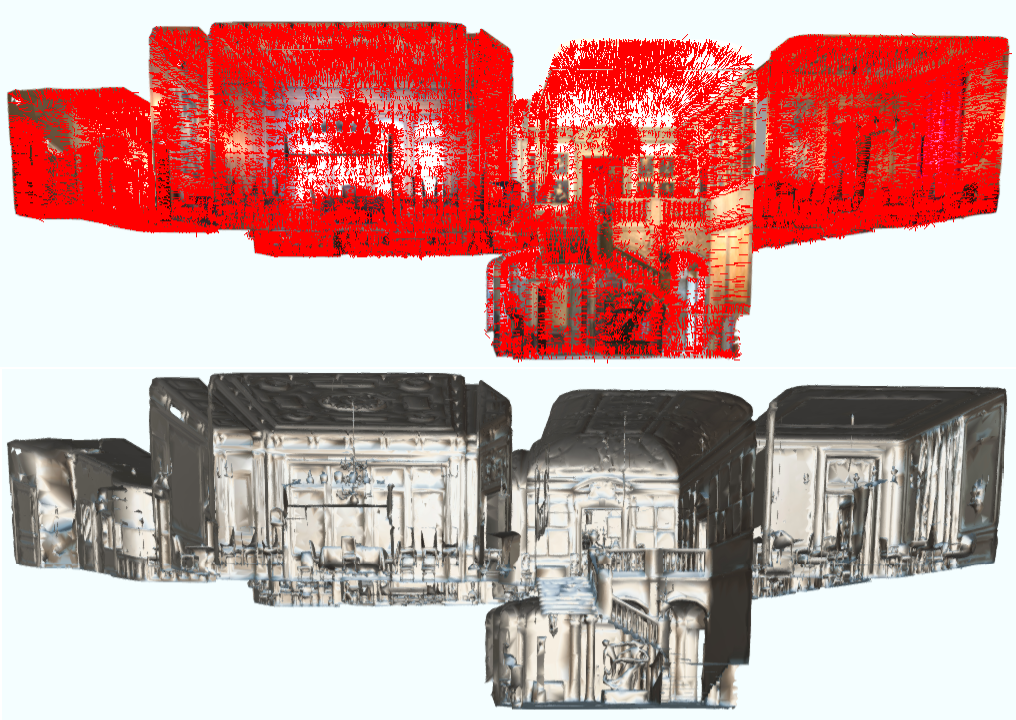}
	}
	\caption{(a) 3D model of the Hallwyl museum; (b) Top: Model vertices to produce Lidar-like dataset, Bottom: Material Capture including lighting and reflections}
	\label{fig:dummy}
	\vspace{-3mm}
\end{figure}

\begin{figure*}[!ht]
	\centering
	\includegraphics[width=.98\linewidth, height=.19\textheight]{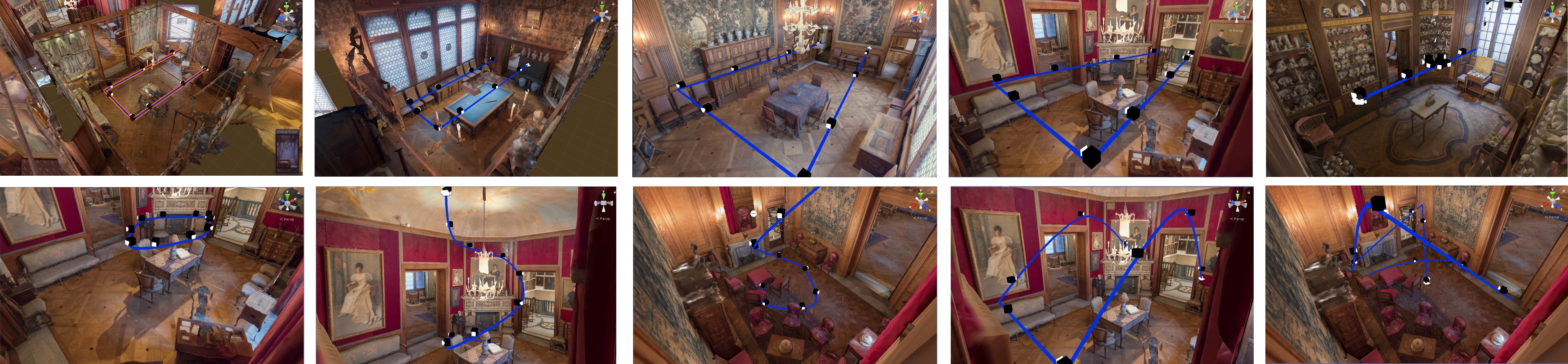}
	\caption{Camera movement trajectory regimes inside different scenes to produce the RGB training images. Styles are: rectangular, spiral, circular, semicircular, and random with forward and backward viewpoints.}
	\label{fig:traj}
	\vspace{-3mm}
\end{figure*}

\begin{figure*}[!ht]
	\centering
	\includegraphics[width=.98\linewidth, height=.28\textheight]{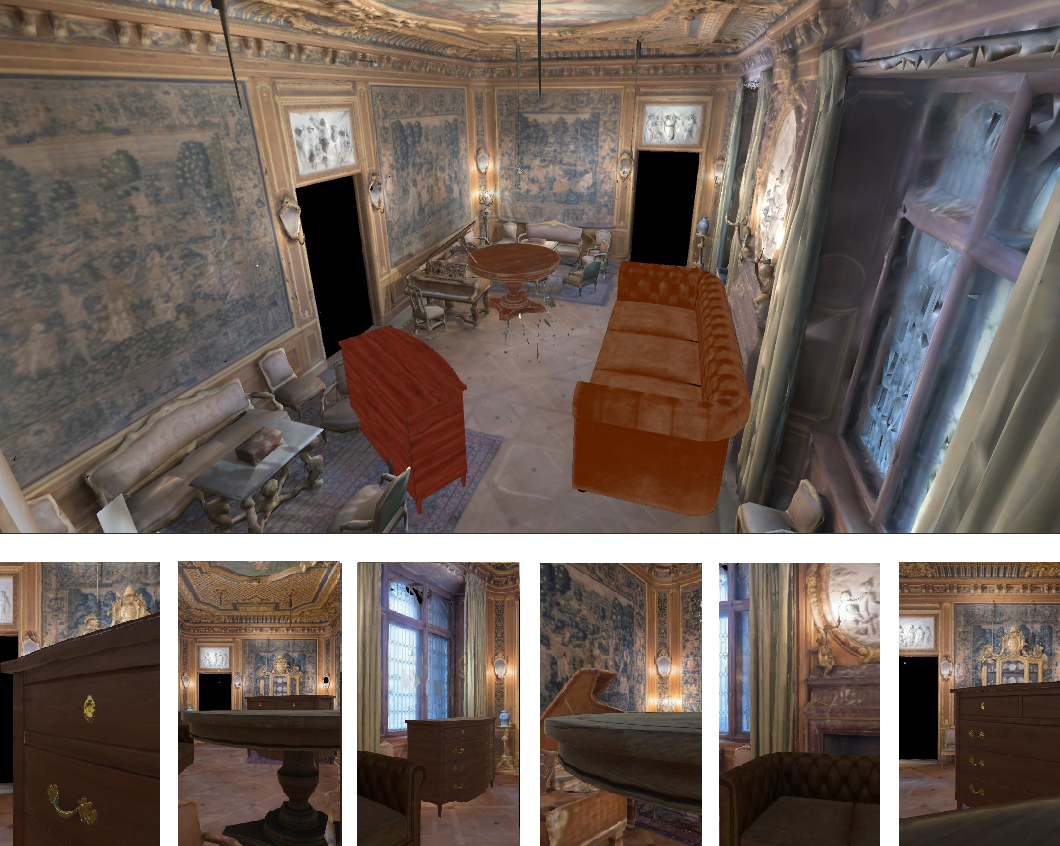}
	\caption{Test samples being created using object insertions which occlude and mask the view patterns, in order to test the model robustness against masking effects. (Top) The big view of the scene, (Bottom) Left-to-right: different camera views and masked patterns due to object occlusions}
	\label{fig:insertion}
%	\vspace{-4mm}
\end{figure*}

%\begin{figure}[!ht]
%	\centering
%	\includegraphics[width=.52\linewidth, height=.47\textheight]{img/classifier}
%	\caption{Scene classifier based on EfficientNet-B0 CNN. \textbf{Input}:RGB data; \textbf{Output}:Associated scene no.}
%	\label{fig:matcap}
%	\vspace{-4mm}
%\end{figure}
\setcounter{footnote}{0} 
The dataset is created in the following procedure. First, 
3D scanned images of the Hallwyl museum in Stockholm\footnote{\url{https://sketchfab.com/TheHallwylMuseum}}, has been sampled using the Unity software, and the normalized outputs are saved.
This museum, as in figure~\ref{fig:3Dmodel1}, has been divided into
9 scenes. More than one Million pure data samples\footnote{\url{http://opensource.alphareality.io}}
has been generated from all scenes using different regimes for
the camera, as depicted in figure~\ref{fig:traj}.
 The equivalent pointcloud data for each of the samples are created as depicted in figure~\ref{fig:3Dmodel_PC}. The complete procedure has been explained in ~\cite{ghofrani2019icps,ghofrani2019lidar}. 
The augmentation process over the images through brightness variations and mask insertions are performed and the outputs are added to both the RGB and pointcloud datasets.
%, as depicted in figures~\ref{fig:brightness}, and figure~\ref{fig:masked}.
In order to evaluate the masking effect on the position regression network, we have created test samples in which some unseen objects are inserted to the scenes, a sample of which is depicted in figure~\ref{fig:insertion}. 
%In the following section, the results of training samples on the test data evaluations are presented.

\vspace{-1mm}
\subsection{Training Of The Model}
\begin{figure*}[!ht]
	\centering
	\includegraphics[width=1\linewidth, height=.185\textheight]{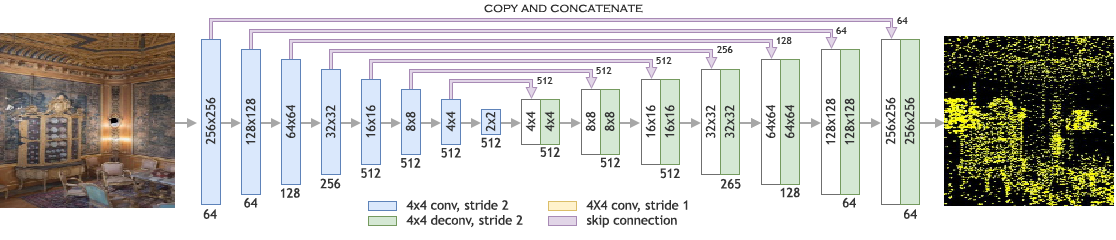}
	\caption{RGB-to-Pointcloud translation, using Pix2Pix GAN}
	\label{fig:unet}
	%	\vspace{-2mm}
\end{figure*}

\vspace{-2mm}
The complete procedure of the model training is shown in algorithm~\ref{al:alg1}. The RGB images created from the segmented scenes, and their corresponding pointcloud (P.C.) images, are the two datasets being available, as discussed in the previous section.
%\vspace{-2mm}
\begin{algorithm}[!ht]
	\caption{Training phase of the Model components}
	\begin{algorithmic} 
		\STATE \textbf{Dataset 1:} RGB images of the scenes.
		\STATE \textbf{Dataset 2:} Point cloud (P.C.) images of the same scenes.
		\STATE \textbf{CNN 1:} Scene classifier (\textbf{EfficientNet B0}) 
		\STATE $\qquad \;$ \textbf{input:} Dataset 1
		\STATE $\qquad \;$ \textbf{Output:} Scene No.
		\STATE \textbf{GAN:} RGB-2-Point Cloud Translator (\textbf{Pix2Pix GAN}) 
		\STATE $\qquad \;$ \textbf{input 1:} Dataset 1
		\STATE $\qquad \;$ \textbf{input 2:} Dataset 2
		\STATE $\qquad \;$ \textbf{Output:} Reconstructed P.C. image $\equiv$ Dataset 3.
		\STATE \textbf{CNN 2:} Multi-Modal Regressor (\textbf{2  EfficientNet CNNs}) 
		\STATE $\qquad \;$ \textbf{input 1:} Dataset 3 
		\STATE $\qquad \;$ \textbf{input 2:} Dataset 1
		\STATE $\qquad \;$ \textbf{Output:} Location.			
	\end{algorithmic}
	\label{al:alg1}
\end{algorithm}
%	\vspace{-2mm}
% These datasets are also freely accessible\footnote{\url{https://www.opensource.alphareality.io}}.
During the test phase, we have no pointcloud data. Therefore, we train a generative neural network based on the UNET architecture~\cite{ronneberger2015u} to estimate the pointcloud equivalent to the input user RGB image. This GAN has a Pix2Pix architecture, as in~\cite{isola2017image}, and has been made robust using augmented data with brightness variations that is depicted in figure~\ref{fig:unet}.
%A sample of such data has been depicted in figure~\ref{fig:rgb2pc}.
%In order to deal with the robustness of the GAN network against brightness conditions of the input images, we have augmented the input samples with different brightness variated samples being assigned to the same pointcloud data samples.
 
%
%One big challenge related to the GAN, is making it robust against different brightness conditions of the input RGB images. To deal with the GAN misperception in this situation, we have augmented the input RGB samples of the GAN with variant brightness samples, which are all assigned to the same point-cloud equivalent image data, as depicted in figure~\ref{fig:brightness}.

\begin{figure}[!ht]
	\centering
	\includegraphics[width=.83\linewidth, height=.63\textheight]{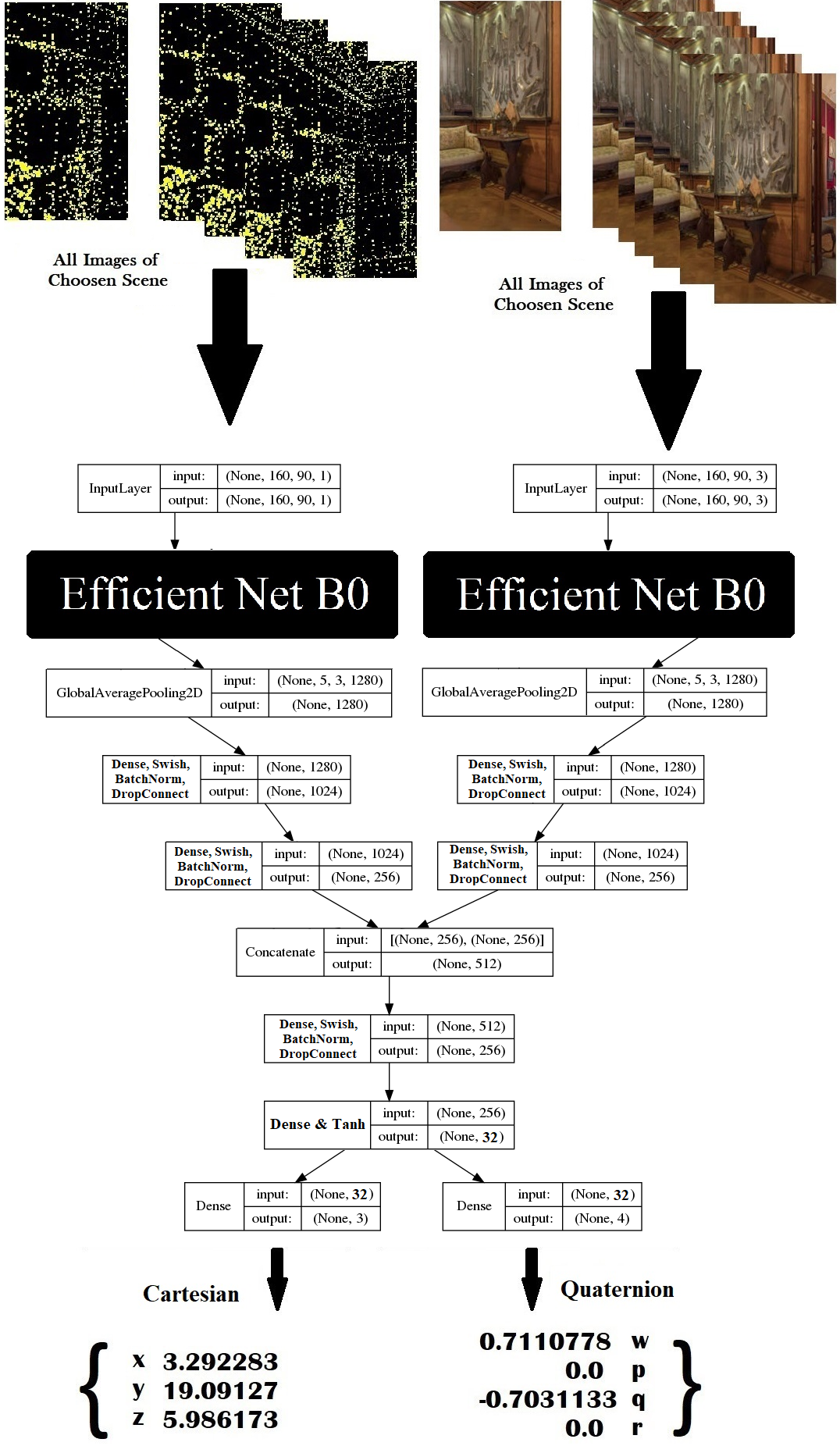}
	\caption{A novel multi-modal regressor CNNs. Left path is only trained on the pointcloud data obtained from the GAN output; Right path only on RGB data.}
	\label{fig:finalRegressor}
	\vspace{-6mm}
\end{figure}

%\begin{figure}[t]
%	\centering
%	\includegraphics[width=1\linewidth, height=.26\textheight]{img/brightness}
%	\caption{Data augmenting by brightness variations w.r.t the reference image: (Top) $50\%$, and $25\%$ light dimming- left to right, (Bottom) $50\%$, and $25\%$ brightening- left to right.}
%	\label{fig:brightness}
%\end{figure}
%\begin{figure}[t]
%	\centering
%	\includegraphics[width=1\linewidth, height=.2\textheight]{img/masked}
%	\caption{Data augmenting by mask insertion and sliding completely over the reference image, on the top-left corner.}
%	\label{fig:masked}
%%	\vspace{-2mm}
%\end{figure}
%For training of the scene classification, a
A CNN based on EfficientNet-B0 architecture~\cite{tan2019efficientnet}, is trained as the scene classifier that gets the input RGB image of a specified scene inside the area of interest and outputs the scene number. The output layers of the standard EfficientNet-B0 is modified according to the problem requirements. Therefore, the stack of MLPs has been converted to $1024 \times 256 \times 32$, and the \textit{Swish} activation function has being involved~\cite{ramachandran2017swish}. In addition, the batch normalization, and dropconnect~\cite{wan2013regularization}, have also been employed in order to avoid model overfitting.

Localizing the user image is a task to be performed by a multi-modal CNN structure, as depicted in figure~\ref{fig:finalRegressor}.
 In this innovative architecture, two individual CNNs are trained, separately. The CNN on the right-hand side of the figure~\ref{fig:finalRegressor}) gets the RGB images and their associated location information as the input-output pairs of samples, according to ICPS-net architecture~\cite{ghofrani2019icps}. The
 left-hand side CNN, takes the reconstructed pointcloud images obtained from the GAN (as the equivalent pointcloud images of the user RGB images) and the corresponding location information, as pairs.
 However, these two structures are both clipped from a middle layer (i.e. the stack of MLP layers) and concatenated, as in figure~\ref{fig:finalRegressor}, to inherit the prominent properties of their ancestors. The trained models of each individual path has been pretrained for 40 epochs~\cite{kumar2017weight}, and after concatenation, the architecture has been again trained completely from the scratch for 90 epochs.

\begin{figure}[!ht]
	\centering
	\includegraphics[width=1\linewidth, height=.2\textheight]{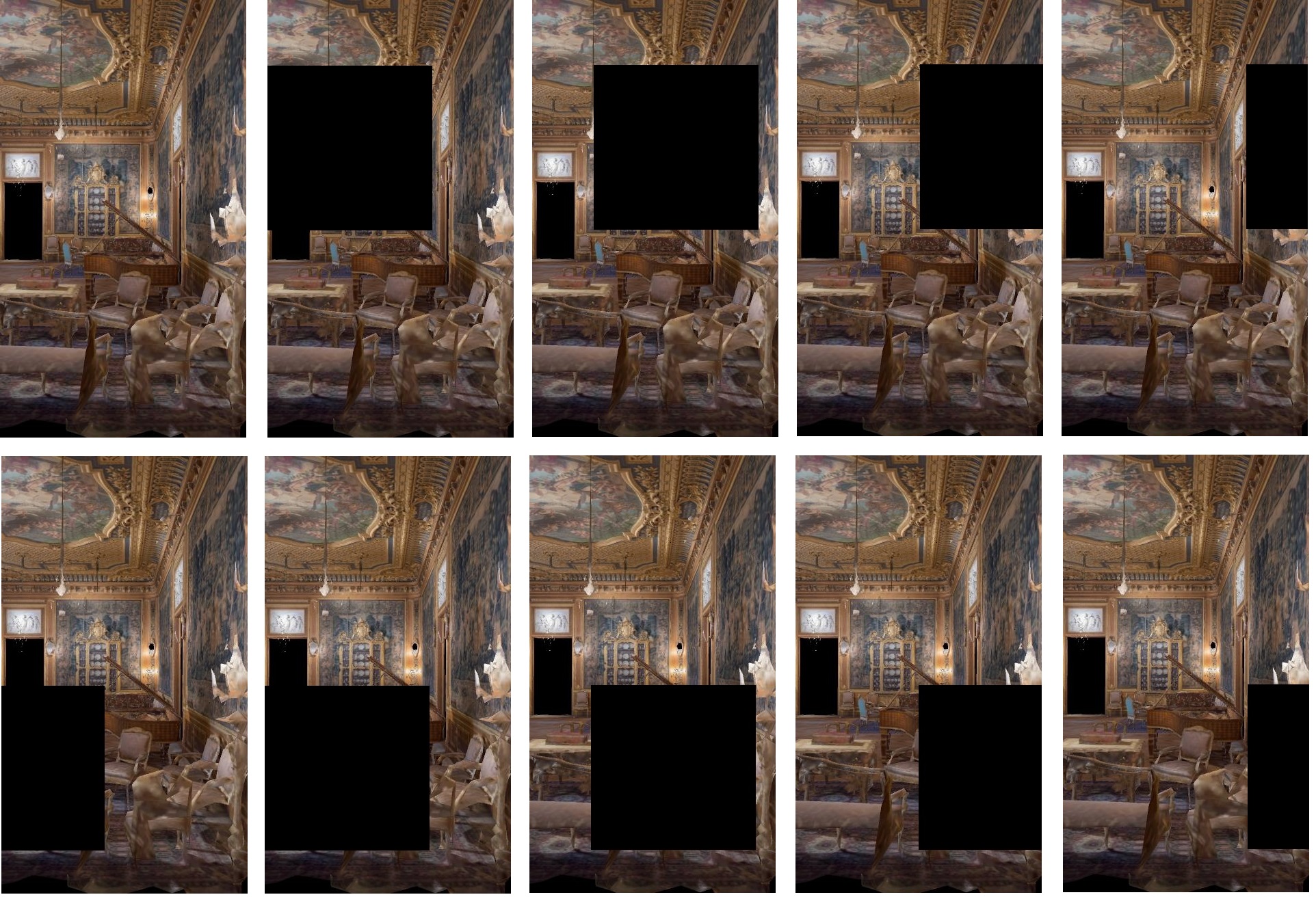}
	\caption{Data augmenting by mask insertion and sliding completely over the reference image, on the top-left corner.}
	\label{fig:masked}
	\vspace{-2mm}
\end{figure}

There are two critical problems for this multimodal regressor. First, each CNN branch contains more than 3 million parameters to be trained. This will lead to model overfitting. We do data augmentation to avoid it. However, the images which are augmented through zooming, sheering, or rotating, can modify the associated position outputs drastically and corrupt the model performance. On the other hand, for indoor areas there are not sufficient indicatives and stationary objects which can perform as the markers of the environment. As a result, moving of the objects may cause serious mismatches between training and test conditions, hence leading to contradictions between what model has been trained with, and what the user feeds to the model. 
In order to solve both problems simultaneously, we used a dark mask window which slides over all images of the input datasets and generates the augmented data for the model, as depicted in figure~\ref{fig:masked}. These masked images correspond to the same output, as the original ones. However, by doing that we can make the model robust against trivial environmental changes. This augmentation is performed in addition to the brightness variations, mentioned earlier. 
%
%%As explained in the training algorithm~\ref{al:alg1}, the GAN network needs an augmented data with the different brightness and masking effects.
%Figure~\ref{fig:brightness}, shows four different brightness levels of changes being applied to a reference image. Accordingly, The upper images show a $50\%$ and $25\%$ dimming of brightness, from left to right respectively, whereas the lower images depict a $50\%$ and $25\%$ increasing of brightness.
The masking augmentation has been applied to the input data including the RGB and associated pointcloud images, as in figure~\ref{fig:masked}. As explained during the model training, the effect of these augmented data would be two-fold: 1) To make the regressors robust against input changes, 2) to avoid the model overfitting.

\begin{figure*}[!ht]
	\centering
	\includegraphics[width=.95\linewidth, height=.325\textheight]{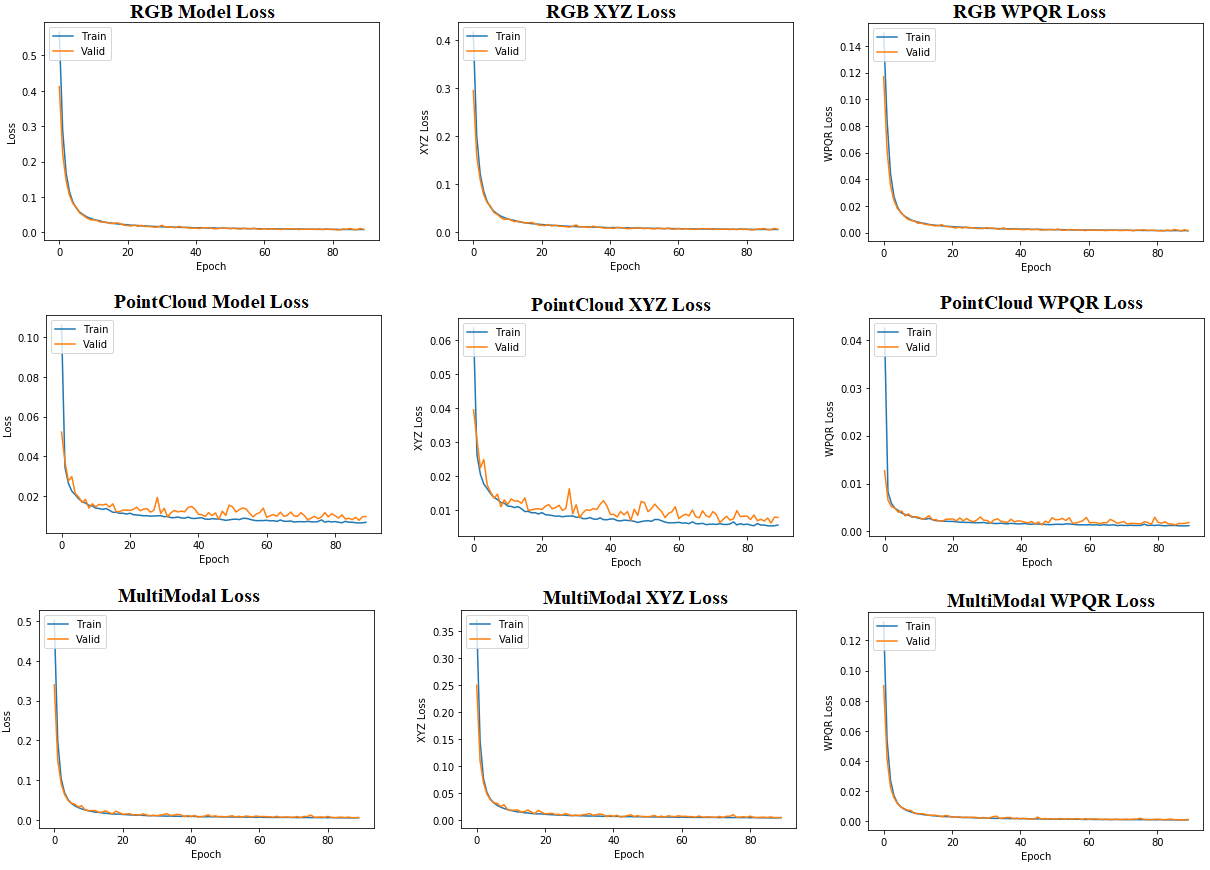}
	\vspace{-1mm}
	\caption{Train vs validation losses of the regressor (Individual CNNs (top 2 rows) vs multi-modal CNN (last row)): (1$^{st}$ row)-left to right: Train vs validation losses of, 1) RGB CNN, 2) Cartesian position est. of the RGB-CNN, 3) quaternion est. of the RGB-CNN;(2$^{nd}$ row)-left to right: Train vs validation losses of, 1) Pointcloud CNN, 2) pointcloud Cartesian est., 3) point-cloud CNN quaternion est., (3$^{rd}$ row)-left to right: Train vs validation losses of, 1) multi-modal CNN, 2) multi-modal Cartesian position est., 3) multi-modal CNN quaternion estimates.}
	\label{fig:RegLoss}
	\vspace{-2mm}
\end{figure*}

The accuracy and loss curves related to the training of the regressor (based on algorithm~\ref{al:alg1}) has been depicted in figure~\ref{fig:RegLoss}.
This figure illustrates the train and validation losses of the regressor, which is a bundle of two CNN architectures.
The first row images, show the training and validation losses of the right-hand side of the regressor structure of figure~\ref{fig:finalRegressor}, which belongs to the RGB images. Model loss, training loss of location information and validation loss of these information are depicted from left-to-right, respectively. Second row shows the same curves for the pointcloud based CNN regressor (left branch of the figure~\ref{fig:finalRegressor}). Third row, further illustrates the aforementioned curves for the integrated regressor model of CNNs. As clearly shown in figure~\ref{fig:RegLoss}, the training and validation losses of the RGB model are pretty well converged. For the pointcloud CNN regressor the convergence is not so good and validation curve is nasty. However, for the combined model, decaying curves mostly follow the RGB model, but the error of training and validation is slightly worse than the RGB-CNN model. Therefore, the premium features of each of the CNN models are inherited in the combined model.

\section{Experimental Analysis and Results}
\vspace{-1mm}
A GTX 1080 NVIDIA GPU, on an Intel 7700 core-i7 CPU, with 32 GBytes of RAM have been employed. Tensorflow 1.13.1 with CUDA 10.1, and Keras
2.2.4 softwares are also used to implement the algorithms.

The RGB sample images are created using a moving camera with an imaging regime and movement styles, as depicted in figure~\ref{fig:traj}. However, the number of the scenes assigned to each interested area is optional. Here, we have divided the Hallwyl 3D graph into 9 different scenes and the RGB images are the ones sampled from each scene, individually. The point-cloud pair of these images are further extracted using a LiDAR system. The created datasets are divided into $60\%$ training, $20\%$ validation, and $20\%$ test data samples. The regression results are further evaluated on the unseen $20\%$ of data samples. The output positioning data has also been normalized to the values between $1$, and $-1$, in order to expedite the convergence process. The normalization is performed as,
%\vspace{-2mm}
\begin{equation}
P = 2\, \frac{P - P_{{Min}}}{P_{{Max}}-P_{{Min}}} -1
\end{equation}
where $P$ is the position, and $P_{{Max}}$ and $P_{{Min}}$ are the maximum and minimum samples of each class, respectively.

\subsection{Testing of The Model}
\vspace{-2mm}
The complete test procedure is depicted in the algorithm~\ref{al:alg2}.
The user RGB image is resized to match the input of the classifier CNN. The EfficientNet B0 classifier CNN, could be easily replaced by the MobileNet V2, and the activation function could be replaced by ReLU in order to reduce the computational burden, to the cost of a trivial growth of the estimation error.
%	\vspace{-3mm}
\begin{algorithm}[!ht]
	\caption{Testing phase of the system}
	\begin{algorithmic} 
		\STATE \textbf{Input:} User RGB image $\equiv$ \textbf{inData}.
		\STATE \textbf{CNN 1:} Scene classifier (\textbf{EfficientNet B0}) 
		\STATE $\qquad \;$ \textbf{input:} \textbf{inData}.
		\STATE $\qquad \;$ \textbf{Output:} Scene No.
		\STATE $\qquad \;$ \textbf{pre-process:} Map the user image to the model image size.
		\STATE \textbf{GAN:} RGB-2-Point Cloud Translator (\textbf{Pix2Pix GAN}) 
		\STATE $\qquad \;$ \textbf{input 1:} \textbf{inData}
		\STATE $\qquad \;$ \textbf{Output:} Reconstructed P.C. image of \textbf{inData}.
		\STATE \textbf{CNN 2:} Multi-Modal Regressors (\textbf{2  EfficientNet CNNs}) 
	%	\vspace{-3mm}
		\STATE $\qquad \;$ \textbf{input 1:} Reconstructed P.C. image of \textbf{inData}. 
		\STATE $\qquad \;$ \textbf{input 2:} \textbf{inData}
		\STATE $\qquad \;$ \textbf{Output:} Location.			
	\end{algorithmic}
	\label{al:alg2}
	\vspace{-1mm}
\end{algorithm}
%	\vspace{-2mm}

\begin{figure}[!ht]
	\centering
	\includegraphics[width=1\linewidth, height=.14\textheight]{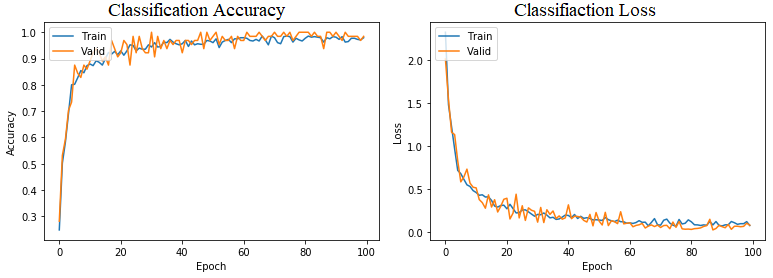}
	\caption{Classification accuracy (left), and Classification loss (right) based on the categorical cross-entropy.}
	\label{fig:classacc_loss}
%		\vspace{-2mm}
\end{figure}

Loss function of the classification CNN is the categorical
cross-entropy, and the model is optimized for maximizing the validation accuracy.
The scene classification accuracy and loss over the training and test data have been depicted in figure~\ref{fig:classacc_loss}, and further shown in table~\ref{tbl:t1}. Using strategies to avoid overfitting such as dropout and dropconnect during training causes the training error exceed the validation error.
%\vspace{-4mm}
\begin{table}[!ht]
	\centering
	\caption{Classifier errors and accuracy values.}
	%	\vspace{-3mm}
	\begin{tabular}{c|c|c|c|}
		\cline{2-4}
		\multicolumn{1}{l|}{}          & Training & Validation & Test     \\ \hline
		\multicolumn{1}{|c|}{Loss}     & 0.06415  & 0.021851   & 0.063899 \\ \hline
		\multicolumn{1}{|c|}{Accuracy} & \%98.514 & \%100      & \%98.099 \\ \hline
	\end{tabular}
	\label{tbl:t1}
	\vspace{-1mm}
\end{table}
%\vspace{-2mm}

\begin{figure}[!ht]
 	\centering
 	\includegraphics[width=.735\linewidth, height=.31\textheight]{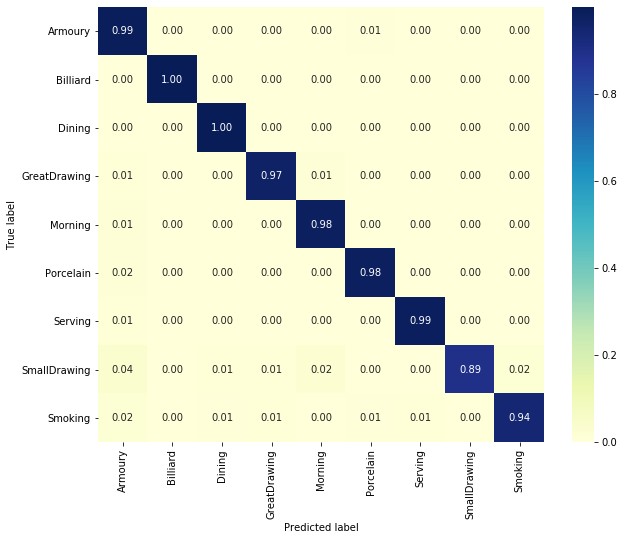}
 	\caption{The confusion matrix for scene classifier using EfficientNet B0 model.}
 	\label{fig:confMatrix}
 %	\vspace{-6mm}
 \end{figure}

  As the confusion matrix in figure~\ref{fig:confMatrix} clarifies the classification has been performed almost perfectly. The worst misclassification belongs to the \textit{Armoury}, and \textit{SmallDrowing} rooms. This is due to the similar patterns which are generated from rectangular trajectory X-Y, in which the camera is close to the wall and the observed patterns get quite similar. This would be completely solved through taking a sequence of panorama-capturing images which create a batch of test images from the same location.
% \vspace{-5mm}
 The regression loss is calculated, as
% \vspace{2mm}
 \begin{equation}
 loss = ||P-\hat{P}||_2 + \frac{1}{\beta} ||\hat{Q}-\frac{Q}{||Q||} ||_2
 \end{equation}
 where $P = [x; y; z]$ is the position data vector, Q is the quaternion, and $\beta$ is the scale factor that makes a balance between position, and the quaternion estimations.
 The regression losses for training, validation, and test set (with $60$,$20$,$20$ partitioning rates, respectively) are shown in table~\ref{tbl:t2}. The comparison among the proposed \textbf{APS} model and the previous models show that the APS model is highly accurate, however it does not carry the problems of the ICPS-net model. The regression loss is chosen based on~\cite{kendall2017geometric,kingma2014adam}.

 In addition, the average errors of the position and quaternion estimates in the output of the proposed model and the competitors are shown in table~\ref{tbl:t3}. The average disruption loss, over the entire test set in a comparison with the other methods is further shown in table~\ref{tbl:t4}, which indicates the superiority of the proposed model.
 \vspace{-5mm}
 \begin{table}[!ht]
	\centering
	\caption{Training, validation, and test losses of the models}
%	\vspace{-3mm}
	\begin{tabular}{|l|l|l|l|}
		\hline
		{\small Method vs Loss} & Training & Validation & Test \\ \hline
		{\small RGB (ICPS-net)} & 0.00781 & 0.00688 & 0.00938 \\ \hline
		{\small P.Cloud (LiDAR ICPS-net)} & 0.01286 & 0.01334 & 0.02964 \\ \hline
		{\small APS} & 0.00813 & 0.00712 & 0.00972 \\ \hline
	\end{tabular}
	\label{tbl:t2}
%	\vspace{-3mm}
\end{table}
\vspace{-10mm}
\begin{table}[!ht]
	\centering
	\caption{Average errors of the methods on the position (X,Y,Z)
		estimates (millimeter), and the Quaternion (degree)}
%	\vspace{-3mm}
	\begin{tabular}{|l|l|l|l|l|}
		\hline
		{\small Method vs AvgErr} & {\scriptsize X} &{\small  Y} & {\small Z} & {\small Quaternion} \\ \hline
		{\small RGB (ICPS-net)} & 2.6 & 3.4 & 1 & 0.0086 \\ \hline
		{\small (LiDAR ICPS-net)} & 19 & 27 & 7.3 & 0.0096\\ \hline
		{\small APS} & 3.1 & 3.9 & 1.3 & 0.0097 \\ \hline
	\end{tabular}
	\label{tbl:t3}
\end{table}
\vspace{-10mm}
\begin{table}[!ht]
	\centering
	\caption{Average disruption error values of the models}
%	\vspace{-3mm}
	\begin{tabular}{|l|c|}
		\hline 
		{\small Method} & {\small Disruption test loss}\tabularnewline
		\hline 
		\hline 
		{\small RGB (ICPS-net)} & 0.0219\tabularnewline
		\hline 
		{\small P.Cloud (LiDAR ICPS-net)} & 0.0365\tabularnewline
		\hline 
		{\small APS} & 0.0167\tabularnewline
		\hline 
	\end{tabular}
	\label{tbl:t4}
%	\vspace{-3mm}
\end{table}
%\vspace{-8mm}

\section{Conclusion}
\vspace{-2mm}
In this paper an end-to-end system has been proposed to address the indoor positioning problem. This work improves the previously proposed systems, namely ICPS-net, and LiDAR ICPS-net. The ICPS-net has become robust against environmental changes due to objects with dynamicity in the background. The LiDAR ICPS-net had the advantage of easier data generation and more robustness against input variations. However, it suffered from lack of precision. Another novelty has been through using the Pix2Pix GAN to generate the point-cloud data which could help data augmentation by generating the images with a distribution close to the dataset. While new data prevents model overfitting, it helps providing robust point-cloud data when input RGB images are masked and occluded. The third novelty has been a multi-modal CNN which merges two individual CNN models and outperforms them for both the regression precision, and the convergence ability. The Cartesian position and quaternion estimates, have been remarkably improved with respect to the SOTA. The novelties of the proposed model simplifies leveraging of the system in various applications, such as large buildings, malls, tunnels, and AR/VR applications. 

\bibliographystyle{splncs04}
\vspace{-2mm}
\bibliography{indoor_ref}

\begin{thebibliography}{10}
\providecommand{\url}[1]{\texttt{#1}}
\providecommand{\urlprefix}{URL }
\providecommand{\doi}[1]{https://doi.org/#1}

\bibitem{brahmbhatt2018geometry}
Brahmbhatt, S., Gu, J., Kim, K., Hays, J., Kautz, J.: Geometry-aware learning
  of maps for camera localization. In: Proceedings of the IEEE Conference on
  Computer Vision and Pattern Recognition. pp. 2616--2625 (2018)

\bibitem{caso2019vifi}
Caso, G., De~Nardis, L., Lemic, F., Handziski, V., Wolisz, A., Di~Benedetto,
  M.G.: Vifi: Virtual fingerprinting wifi-based indoor positioning via
  multi-wall multi-floor propagation model. IEEE Transactions on Mobile
  Computing  (2019)

\bibitem{duque2017improved}
Duque~Domingo, J., Cerrada, C., Valero, E., Cerrada, J.: An improved indoor
  positioning system using rgb-d cameras and wireless networks for use in
  complex environments. Sensors  \textbf{17}(10), ~2391 (2017)

\bibitem{duque2016indoor}
Duque~Domingo, J., Cerrada, C., Valero, E., Cerrada, J.A.: Indoor positioning
  system using depth maps and wireless networks. Journal of Sensors
  \textbf{2016} (2016)

\bibitem{faragher2014analysis}
Faragher, R., Harle, R.: An analysis of the accuracy of bluetooth low energy
  for indoor positioning applications. In: Proceedings of the 27th
  International Technical Meeting of The Satellite Division of the Institute of
  Navigation (ION GNSS+ 2014). vol.~812, pp. 201--210 (2014)

\bibitem{ghofrani2019icps}
Ghofrani, A., Toroghi, R.M., Tabatabaie, S.M.: Icps-net: An end-to-end
  rgb-based indoor camera positioning system using deep convolutional neural
  networks. arXiv preprint arXiv:1910.06219  (2019)

\bibitem{ghofrani2019lidar}
Ghofrani, A., Toroghi, R.M., Tabatabaie, S.M., Tabasi, S.M.: Lidar icps-net:
  Indoor camera positioning based-on generative adversarial network for rgb to
  point-cloud translation. arXiv preprint arXiv:1911.05871  (2019)

\bibitem{guan2016vision}
Guan, K., Ma, L., Tan, X., Guo, S.: Vision-based indoor localization approach
  based on surf and landmark. In: 2016 International Wireless Communications
  and Mobile Computing Conference (IWCMC). pp. 655--659. IEEE (2016)

\bibitem{henriques2018mapnet}
Henriques, J.F., Vedaldi, A.: Mapnet: An allocentric spatial memory for mapping
  environments. In: proceedings of the IEEE Conference on Computer Vision and
  Pattern Recognition. pp. 8476--8484 (2018)

\bibitem{henry2014rgb}
Henry, P., Krainin, M., Herbst, E., Ren, X., Fox, D.: Rgb-d mapping: Using
  depth cameras for dense 3d modeling of indoor environments. In: Experimental
  robotics. pp. 477--491. Springer (2014)

\bibitem{huang2019hybrid}
Huang, K., He, K., Du, X.: A hybrid method to improve the ble-based indoor
  positioning in a dense bluetooth environment. Sensors  \textbf{19}(2), ~424
  (2019)

\bibitem{isola2017image}
Isola, P., Zhu, J.Y., Zhou, T., Efros, A.A.: Image-to-image translation with
  conditional adversarial networks. In: Proceedings of the IEEE conference on
  computer vision and pattern recognition. pp. 1125--1134 (2017)

\bibitem{jianyong2014rssi}
Jianyong, Z., Haiyong, L., Zili, C., Zhaohui, L.: Rssi based bluetooth low
  energy indoor positioning. In: 2014 International Conference on Indoor
  Positioning and Indoor Navigation (IPIN). pp. 526--533. IEEE (2014)

\bibitem{kendall2017geometric}
Kendall, A., Cipolla, R.: Geometric loss functions for camera pose regression
  with deep learning. In: Proceedings of the IEEE Conference on Computer Vision
  and Pattern Recognition. pp. 5974--5983 (2017)

\bibitem{kendall2015posenet}
Kendall, A., Grimes, M., Cipolla, R.: Posenet: A convolutional network for
  real-time 6-dof camera relocalization. In: Proceedings of the IEEE
  international conference on computer vision. pp. 2938--2946 (2015)

\bibitem{kingma2014adam}
Kingma, D.P., Ba, J.: Adam: A method for stochastic optimization. arXiv
  preprint arXiv:1412.6980  (2014)

\bibitem{kumar2017weight}
Kumar, S.K.: On weight initialization in deep neural networks. arXiv preprint
  arXiv:1704.08863  (2017)

\bibitem{lai2018development}
Lai, C.C., Su, K.L.: Development of an intelligent mobile robot localization
  system using kinect rgb-d mapping and neural network. Computers \& Electrical
  Engineering  \textbf{67},  620--628 (2018)

\bibitem{liang2013image}
Liang, J.Z., Corso, N., Turner, E., Zakhor, A.: Image based localization in
  indoor environments. In: 2013 Fourth International Conference on Computing
  for Geospatial Research and Application. pp. 70--75. IEEE (2013)

\bibitem{liang2015image}
Liang, J.Z., Corso, N., Turner, E., Zakhor, A.: Image-based positioning of
  mobile devices in indoor environments. In: Multimodal Location Estimation of
  Videos and Images, pp. 85--99. Springer (2015)

\bibitem{lin2015mobile}
Lin, X.Y., Ho, T.W., Fang, C.C., Yen, Z.S., Yang, B.J., Lai, F.: A mobile
  indoor positioning system based on ibeacon technology. In: 2015 37th Annual
  International Conference of the IEEE Engineering in Medicine and Biology
  Society (EMBC). pp. 4970--4973. IEEE (2015)

\bibitem{liu2019accurate}
Liu, T., Zhang, X., Li, Q., Fang, Z., Tahir, N.: An accurate visual-inertial
  integrated geo-tagging method for crowdsourcing-based indoor localization.
  Remote Sensing  \textbf{11}(16), ~1912 (2019)

\bibitem{ma2019wifi}
Ma, Z., Wu, B., Poslad, S.: A wifi rssi ranking fingerprint positioning system
  and its application to indoor activities of daily living recognition.
  International Journal of Distributed Sensor Networks  \textbf{15}(4),
  1550147719837916 (2019)

\bibitem{mekki2019indoor}
Mekki, K., Bajic, E., Meyer, F.: Indoor positioning system for iot device based
  on ble technology and mqtt protocol. In: 2019 IEEE 5th World Forum on
  Internet of Things (WF-IoT). pp. 787--792. IEEE (2019)

\bibitem{mendoza2019meta}
Mendoza-Silva, G.M., Torres-Sospedra, J., Huerta, J.: A meta-review of indoor
  positioning systems. Sensors  \textbf{19}(20), ~4507 (2019)

\bibitem{morgado2019beacons}
Morgado, F., Martins, P., Caldeira, F.: Beacons positioning detection, a novel
  approach. In: The 10th International Conference on Ambient Systems, Networks
  and Technologies (ANT 2019. vol.~151, pp. 23--30. Elsevier (2019)

\bibitem{ramachandran2017swish}
Ramachandran, P., Zoph, B., Le, Q.V.: Swish: a self-gated activation function.
  arXiv preprint arXiv:1710.05941  \textbf{7} (2017)

\bibitem{ronneberger2015u}
Ronneberger, O., Fischer, P., Brox, T.: U-net: Convolutional networks for
  biomedical image segmentation. In: International Conference on Medical image
  computing and computer-assisted intervention. pp. 234--241. Springer (2015)

\bibitem{russ2019augmented}
Russ, M., Angermayer, L., Pichler, G., Koller, S., Kiss, E., Mitterhoefer, C.,
  Lasser, M., Lerchbaumer, J.I., Wagner, P.: Augmented reality systems and
  methods for providing player action recommendations in real time (Feb~14
  2019), uS Patent App. 15/852,088

\bibitem{sandler2018mobilenetv2}
Sandler, M., Howard, A., Zhu, M., Zhmoginov, A., Chen, L.C.: Mobilenetv2:
  Inverted residuals and linear bottlenecks. In: Proceedings of the IEEE
  Conference on Computer Vision and Pattern Recognition. pp. 4510--4520 (2018)

\bibitem{sattler2012improving}
Sattler, T., Leibe, B., Kobbelt, L.: Improving image-based localization by
  active correspondence search. In: European conference on computer vision. pp.
  752--765. Springer (2012)

\bibitem{sattler2016efficient}
Sattler, T., Leibe, B., Kobbelt, L.: Efficient \& effective prioritized
  matching for large-scale image-based localization. IEEE transactions on
  pattern analysis and machine intelligence  \textbf{39}(9),  1744--1756 (2016)

\bibitem{shao2018ibeacon}
Shao, S., Shuo, N., Kubota, N.: An ibeacon indoor positioning system based on
  multi-sensor fusion. In: 2018 Joint 10th International Conference on Soft
  Computing and Intelligent Systems (SCIS) and 19th International Symposium on
  Advanced Intelligent Systems (ISIS). pp. 1115--1120. IEEE (2018)

\bibitem{sharp2019indoor}
Sharp, I., Yu, K.: Indoor wifi positioning. In: Wireless Positioning:
  Principles and Practice, pp. 219--240. Springer (2019)

\bibitem{shen2019low}
Shen, Z., Liu, J., Zheng, Y., Cao, L.: A low-cost mobile vr walkthrough system
  for displaying multimedia works based on unity3d. In: 2019 14th International
  Conference on Computer Science \& Education (ICCSE). pp. 415--419. IEEE
  (2019)

\bibitem{sieberth2019applying}
Sieberth, T., Dobay, A., Affolter, R., Ebert, L.C.: Applying virtual reality in
  forensics--a virtual scene walkthrough. Forensic Science, Medicine and
  Pathology  \textbf{15}(1),  41--47 (2019)

\bibitem{tan2019efficientnet}
Tan, M., Le, Q.V.: Efficientnet: Rethinking model scaling for convolutional
  neural networks. arXiv preprint arXiv:1905.11946  (2019)

\bibitem{valgren2010sift}
Valgren, C., Lilienthal, A.J.: Sift, surf \& seasons: Appearance-based
  long-term localization in outdoor environments. Robotics and Autonomous
  Systems  \textbf{58}(2),  149--156 (2010)

\bibitem{wan2013regularization}
Wan, L., Zeiler, M., Zhang, S., Le~Cun, Y., Fergus, R.: Regularization of
  neural networks using dropconnect. In: International conference on machine
  learning. pp. 1058--1066 (2013)

\bibitem{wang2019high}
Wang, R., Wan, W., Di, K., Chen, R., Feng, X.: A high-accuracy
  indoor-positioning method with automated rgb-d image database construction.
  Remote Sensing  \textbf{11}(21), ~2572 (2019)

\bibitem{yang2015wifi}
Yang, C., Shao, H.R.: Wifi-based indoor positioning. IEEE Communications
  Magazine  \textbf{53}(3),  150--157 (2015)

\bibitem{yuan2016rgb}
Yuan, W., Li, Z., Su, C.Y.: Rgb-d sensor-based visual slam for localization and
  navigation of indoor mobile robot. In: 2016 International Conference on
  Advanced Robotics and Mechatronics (ICARM). pp. 82--87. IEEE (2016)

\bibitem{zhang2018real}
Zhang, F., Lei, T., Li, J., Cai, X., Shao, X., Chang, J., Tian, F.: Real-time
  calibration and registration method for indoor scene with joint depth and
  color camera. International Journal of Pattern Recognition and Artificial
  Intelligence  \textbf{32}(07),  1854021 (2018)

\bibitem{zuo2018indoor}
Zuo, Z., Liu, L., Zhang, L., Fang, Y.: Indoor positioning based on bluetooth
  low-energy beacons adopting graph optimization. Sensors  \textbf{18}(11),
  ~3736 (2018)

\end{thebibliography}
\end{document}